\documentclass[11pt]{article}

\usepackage[utf8]{inputenc}
\usepackage[T1]{fontenc}
\usepackage{lmodern}
\usepackage[margin=1in]{geometry}
\usepackage{graphicx}
\usepackage{tikz}
\usetikzlibrary{arrows.meta,shapes.geometric,positioning}
\usepackage[ruled,vlined,linesnumbered]{algorithm2e}
\usepackage{booktabs}
\usepackage{amsmath,amssymb}
\usepackage{microtype}
\usepackage{xcolor}
\usepackage[numbers,sort&compress]{natbib}
\usepackage{authblk}
\usepackage[colorlinks=true,linkcolor=blue,citecolor=blue,urlcolor=blue]{hyperref}
\usepackage{fancyhdr}
\title{\textbf{Scale and Selection: What Makes Automatic Harness Evolution Work for Visual-Interface Robot Agents}}

\author{Zhijie Wei\textsuperscript{1}, Ferris Tan\textsuperscript{2}, Jinghui Wang\textsuperscript{1,*}, Hao Zhou\textsuperscript{2}}
\date{}

\begin{document}

\maketitle
\begingroup
\renewcommand{\thefootnote}{\arabic{footnote}}
\footnotetext[1]{NovaXBot}
\footnotetext[2]{QuantGroup}
\renewcommand{\thefootnote}{\fnsymbol{footnote}}
\footnotetext[1]{Corresponding author.}
\endgroup

\begin{abstract}
When an off-the-shelf coding agent is used directly as a robot policy, observing a browser-based 3D interface through screenshots and acting by posing a virtual target gripper through a few tools, the agent's harness, its prompts, tools, and control rules, largely determines success, and until now it has been written by hand. We show that this harness can be improved automatically by another coding agent, the optimizer agent, and report two findings about what makes it work. First, the number of rollouts the optimizer agent sees per round governs whether the evolved harness generalizes: with rounds fixed, held-out success rises as the training batch grows from 5 to 100 rollouts, while small batches overfit the training tasks. Second, the optimizer agent must not be given free rein. Accepting every revision it proposes degrades the harness within a few rounds; a simple Champion--Challenger rule that promotes a revision only if it beats the incumbent on the same fixed cases turns the same loop into one that raises held-out success on LIBERO tasks from 51\% to 67\% over 30 rounds. The evolved harness is also a natural place to attach a pretrained policy, through one tool that hands it control. Writing the rollouts of that combination back into the policy, by fine-tuning domain experts on them and merging the experts into one policy with multi-teacher on-policy distillation, yields a standalone policy at 62.2\% on RoboCasa365 under the official protocol, 4.8 points above the policy it started from with no harness at inference; the evolved harness on top of that distilled policy reaches 71.2\% on the RoboCasa365 leaderboard benchmark, 13.1 points above the best published entry, and the loop closes when harness evolution resumes on top of the stronger policy. Automatic harness evolution for visual-interface robot agents is thus feasible, but its gains hinge on the rollout scale behind each decision and on how the optimizer agent's revisions are selected.
\end{abstract}

\section{Introduction}

Foundation-model coding agents such as Codex and Claude Code can now control robots without any robot-specific training \citep{via2026,rho2026,guava2026}, and since the release of GPT-6 Astra, frontier models have been evaluated directly as robot policies across manipulation, dexterous hands, navigation, and humanoid control \citep{galbot2026}. \citet{via2026} present the robot as a \emph{visual interface}, a browser-based 3D point-cloud GUI with a virtual target gripper and a few MCP tools, and let the agent operate it as a person would operate 3D design software, reaching 96.7\% on three LIBERO-Goal tasks. We refer to such a system as a \emph{visual-interface robot agent}: a coding agent that controls a robot by operating a visual interface. What the agent can do, however, depends heavily on its \emph{harness}: the prompt, tool interface, observation format, control protocol, and verification rules that sit between model and robot \citep{metaharness2026,selfharness2026}. In that system, a single textual waypoint demonstration in the prompt lifts one agent from 70\% to 100\% while leaving another unchanged \citep{via2026}.

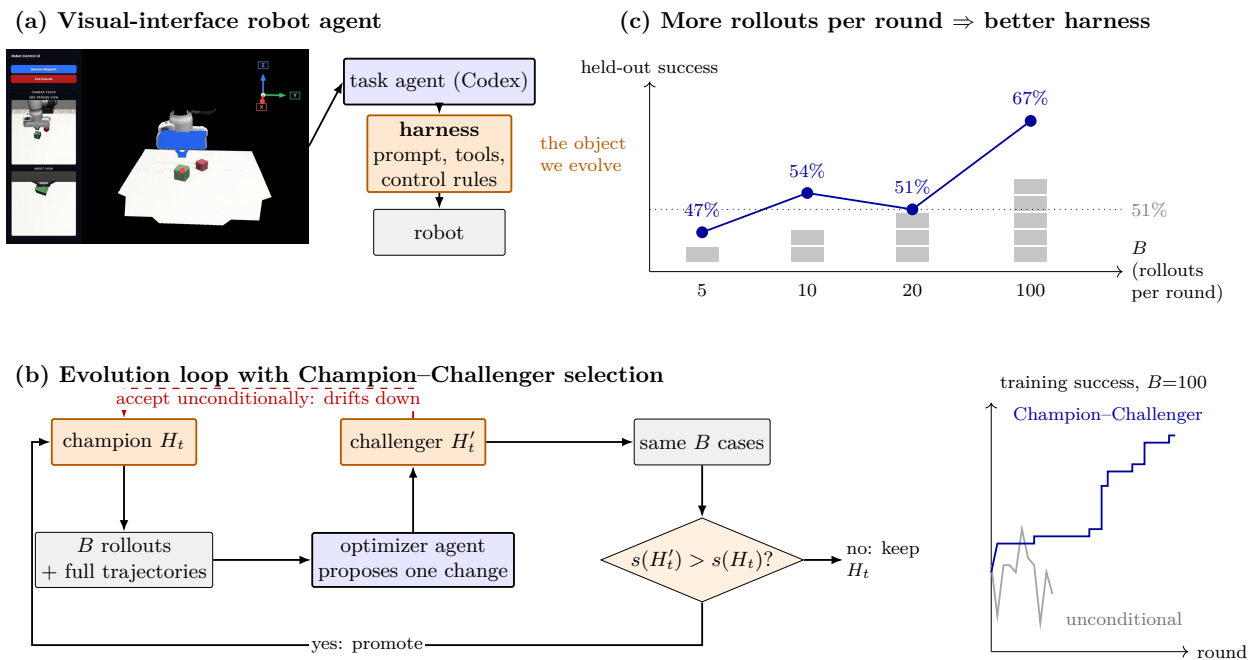
\begin{figure}[t]
\centering
\resizebox{\textwidth}{!}{%
\begin{tikzpicture}[
  x=1mm, y=1mm,
  font=\footnotesize,
  box/.style={draw, rounded corners=1pt, align=center, inner sep=2.5pt, minimum height=7mm, font=\footnotesize},
  agent/.style={box, fill=blue!10, thick},
  data/.style={box, fill=gray!12},
  hl/.style={box, fill=orange!18, thick, draw=orange!70!black},
  dec/.style={draw, diamond, aspect=2.4, inner sep=0.5pt, fill=orange!12, font=\footnotesize},
  arr/.style={-{Latex[length=1.8mm]}, thick},
  lbl/.style={font=\scriptsize, fill=white, inner sep=1pt},
  note/.style={font=\scriptsize, text=gray, align=left},
  ptitle/.style={font=\small\bfseries, anchor=west}
]
\node[ptitle] at (0,104) {(a) Visual-interface robot agent};
\node[anchor=north west, inner sep=0pt] (img) at (0,100) {%
  \IfFileExists{via_ui.jpg}{\includegraphics[width=46mm]{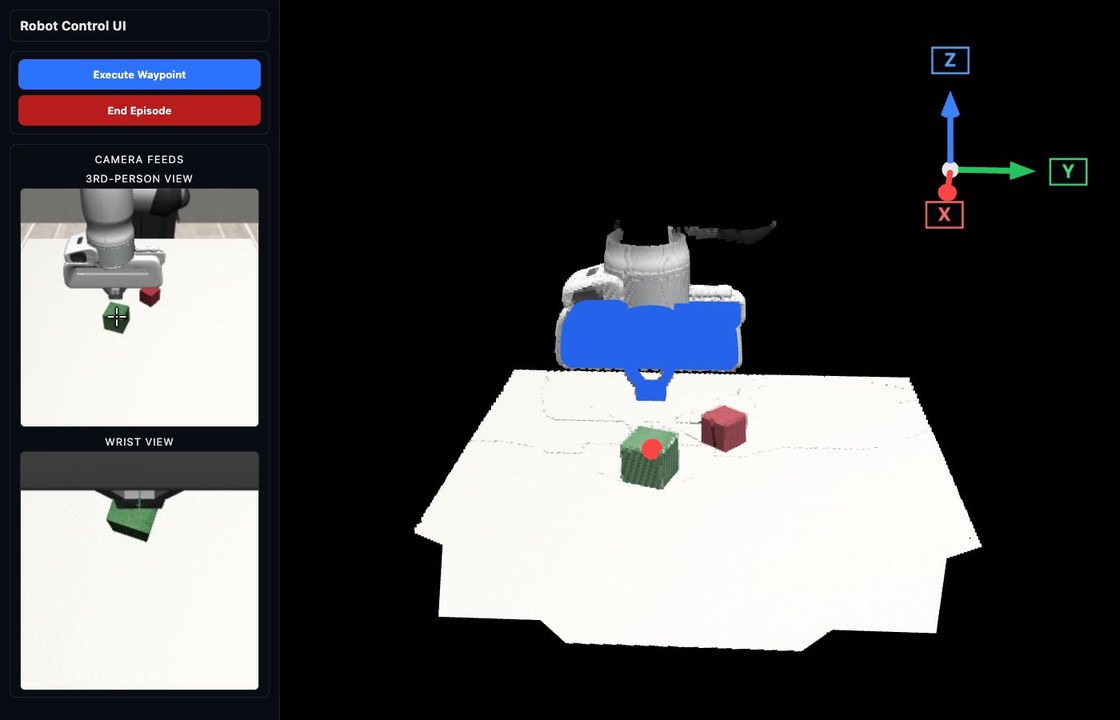}}{\tikz\node[draw, fill=black!80, text=white, minimum width=46mm, minimum height=29.6mm, align=center]{visual interface\\(screenshot)};}};
\node[agent, minimum width=20mm] (ta) at (66,95) {task agent (Codex)};
\node[hl, minimum width=20mm] (hn) at (66,84) {\textbf{harness}\\prompt, tools,\\control rules};
\node[data, minimum width=20mm] (rb) at (66,72) {robot};
\draw[arr] (ta) -- (hn);
\draw[arr] (hn) -- (rb);
\draw[arr] (img.east) -- (ta.west);
\node[font=\scriptsize, text=orange!70!black, anchor=west, align=left] at (80,84) {the object\\we evolve};

\node[ptitle] at (92,104) {(c) More rollouts per round $\Rightarrow$ better harness};
\draw[->] (98,66) -- (98,95) node[anchor=south, font=\scriptsize] {held-out success};
\draw[->] (98,66) -- (170,66) node[anchor=west, font=\scriptsize, align=left] {$B$\\(rollouts\\per round)};
\foreach \x/\n/\b in {106/1/5, 122/2/10, 138/3/20, 156/5/100} {
  \foreach \i in {1,...,\n} { \fill[gray!45] (\x-2.5, 67.5+2.6*\i-2.6) rectangle ++(5,2.2); }
  \node[font=\scriptsize, anchor=north] at (\x,65.5) {\b};
}
\draw[thick, blue!60!black] (106,72) -- (122,78) -- (138,75.5) -- (156,89);
\foreach \x/\y/\v in {106/72/47, 122/78/54, 138/75.5/51, 156/89/67} {
  \fill[blue!60!black] (\x,\y) circle (0.9);
  \node[font=\scriptsize, anchor=south, blue!60!black] at (\x,\y+1) {\v\%};
}
\draw[dotted] (98,75.5) -- (170,75.5) node[anchor=west, font=\scriptsize, text=gray] {51\%};

\node[ptitle] at (0,50) {(b) Evolution loop with Champion--Challenger selection};
\node[hl, minimum width=22mm] (ch) at (18,40) {champion $H_t$};
\node[data, minimum width=22mm] (ro) at (18,22) {$B$ rollouts\\+ full trajectories};
\node[agent, minimum width=22mm] (op) at (62,22) {optimizer agent\\proposes one change};
\node[hl, minimum width=22mm] (cl) at (62,40) {challenger $H'_t$};
\node[data, minimum width=20mm] (ro2) at (106,40) {same $B$ cases};
\node[dec] (dc) at (106,22) {$s(H'_t)>s(H_t)$?};
\draw[arr] (ch) -- (ro);
\draw[arr] (ro) -- (op);
\draw[arr] (op) -- (cl);
\draw[arr] (cl) -- (ro2);
\draw[arr] (ro2) -- (dc);
\draw[arr] (dc.south) -- (106,9) -- (4,9) -- (4,40) -- (ch.west);
\node[lbl] at (55,9) {yes: promote};
\draw[arr] (dc.east) -- ++(6,0) node[lbl, anchor=west, align=left] {no: keep\\$H_t$};
\draw[arr, red!70!black, dashed] (cl.north) -- ++(0,4.5) -| (ch.north);
\node[lbl, text=red!70!black] at (40,46.5) {accept unconditionally: drifts down};
\node[font=\scriptsize, anchor=west] at (150,49) {training success, $B{=}100$};
\draw[->] (150,8) -- (150,46);
\draw[->] (150,8) -- (180,8) node[anchor=west, font=\scriptsize] {round};
\draw[gray!70, thick] plot coordinates {(150,21.2) (150.93,13.5) (151.87,21.2) (152.8,21.2) (153.73,20.1) (154.67,26.7) (155.6,22.3) (156.53,21.2) (157.47,12.4) (158.4,20.1) (159.33,16.8)};
\draw[blue!60!black, thick] plot coordinates {(150,20.1) (150.93,24.5) (156.53,24.5) (156.53,25.6) (164.93,25.6) (164.93,26.7) (166.8,26.7) (166.8,33.3) (167.73,33.3) (167.73,35.5) (171.47,35.5) (171.47,36.6) (173.33,36.6) (173.33,39.9) (177.07,39.9) (177.07,41) (178,41)};
\node[font=\scriptsize, text=gray!90!black, anchor=west] at (160,13) {unconditional};
\node[font=\scriptsize, text=blue!60!black, anchor=west] at (152,44) {Champion--Challenger};
\end{tikzpicture}%
}
\caption{Overview. (a) The task agent is an unmodified coding agent that controls a robot through a visual interface; the agent's harness is what we evolve. (b) Each round, an optimizer agent reads the full trajectories of $B$ rollouts and proposes one bounded revision; under Champion--Challenger selection the revision is promoted only if it strictly beats the incumbent on the same fixed cases, whereas accepting every revision drifts downward within ten rounds (inset: training success at $B=100$). (c) Held-out success of the evolved harness rises with the number of rollouts per round, from 47\% at $B=5$ to 67\% at $B=100$.}
\label{fig:overview}
\end{figure}

This harness was written and tuned by hand. We show that it can instead be improved automatically by another coding agent. We call the agent that runs the robot the \emph{task agent} and the agent that revises its harness the \emph{optimizer agent} (Figure~\ref{fig:overview}). Each round, the optimizer agent reads evidence from a batch of task-agent rollouts, proposes one bounded change to the harness, and the revised harness is evaluated again; the underlying models are never trained. Automatic harness optimization has been studied for software agents \citep{metaharness2026,selfharness2026}, where a rollout takes seconds and is graded by a unit test. For a visual-interface robot agent, each rollout spans tens to over a hundred model turns, each consuming fresh screenshots of the interface, and yields a single noisy binary outcome. This changes what the loop needs in order to work, and the rest of this paper is about two such requirements.

The first requirement is rollout scale. Whether a proposed revision is promoted depends on comparing two success counts on the training sample, and each count is a sum of noisy binary outcomes. With five rollouts per round, one flipped outcome moves the success rate by twenty points, so a revision can be promoted, and the next round built on it, purely by luck; the optimizer agent also sees too few failures to distinguish a systematic mechanism from an accident. Enlarging the batch raises the signal-to-noise ratio of every promotion decision and broadens the evidence the optimizer agent reasons from. Holding the number of rounds fixed at thirty and growing the training set from 5 to 10, 20, and 100 rollouts, held-out success rises from 47\% to 67\%. Small training sets do not merely help less: at ten rollouts the loop reaches 70\% on the training set but only 54\% held-out, having fitted the particular cases it saw. A trustworthy, generalizing, steadily improving harness therefore requires rollout scale behind each decision.

The second requirement is that the optimizer agent must not be given free rein. The simplest loop accepts every revision it proposes, so the harness follows a single chain of commits. With expensive, noisy rollouts it fails: performance rises for a few rounds and then drifts downward, falling from 42\% to 34\% on the training set within ten rounds, because a revision that scored well on a lucky batch is kept and every later revision builds on it. The most basic safeguard repairs this: \emph{Champion--Challenger} selection, in which a revision is promoted only if it strictly beats the incumbent harness on the same fixed evaluation set and is otherwise logged and discarded. With this rule and a training set of 100 rollouts, optimizing on tasks drawn from robosuite and LIBERO \citep{libero2023} and reporting on a disjoint held-out set of 100 rollouts from different tasks, thirty rounds raise held-out success from 51\% to 67\%.

The evolved harness is also a natural place to attach a learned policy. Adding one tool that hands control to the Xiaomi-Robotics-1 vision-language-action policy \citep{xiaomi2026}, with the agent deciding when to invoke it and when to take over, turns the harness into a data engine: its rollouts, written back into the policy as domain experts and merged by multi-teacher on-policy distillation, yield a standalone policy at 62.2\% under the official RoboCasa365 protocol, 4.8 points above the policy alone with no harness at inference (Section~\ref{sec:exp-distill}). The evolved harness on top of that distilled policy reaches 71.2\% on the RoboCasa365 leaderboard benchmark, 13.1 points above the best published entry and 13.8 points above the original policy alone (Section~\ref{sec:exp-combo}).

Our contributions are:
\begin{enumerate}
  \item \textbf{The first scaling study of harness evolution for a visual-interface robot agent, showing that rollout scale is a necessary condition.} An optimizer agent improves the hand-written harness with no training of either model, but only with enough rollouts behind each decision: with rounds fixed, held-out success rises as the training batch grows from 5 to 100 rollouts, reaching 67\% against 51\% for the baseline, while small batches overfit.
  \item \textbf{The finding that the optimizer agent must not be given free rein.} We show that unconditional acceptance of its revisions degrades performance within a few rounds, and that Champion--Challenger selection, the most basic safeguard, turns the same loop into one that improves over 30 rounds.
  \item \textbf{A route from the evolved harness back into the policy.} The rollouts produced by the evolved harness are written back into the policy as domain experts, and the experts are merged into one policy by multi-teacher on-policy distillation (MOPD). The resulting standalone policy reaches 62.2\% under the official protocol, 4.8 points above the policy it started from, with no harness at inference, and closes a loop in which the stronger policy becomes the executor for the next round of harness evolution.
  \item \textbf{A new best result on the RoboCasa365 leaderboard benchmark.} The evolved harness on top of the policy distilled from its own rollouts, through one tool that hands the policy control, reaches 71.2\%, 13.1 points above the best published entry and 13.8 points above the original policy alone.
\end{enumerate}

\section{Related Work}

\paragraph{Foundation-model agents for robot control.}
Three routes turn foundation models into robot controllers. The dominant one fine-tunes a model into a vision-language-action (VLA) policy on robot trajectories \citep{rt2_2023,openvla2024,pi0_2025}; inference is fast, but the approach needs large robot datasets and yields models smaller and less general than the frontier models it starts from. A second route keeps the model frozen and has it compose perception and control primitives in code \citep{cap2023}, and CaP-X shows that coding-agent success depends strongly on the abstraction level of those primitives \citep{capx2026}. A third lets the model produce intermediate spatial targets, such as value maps or keypoint constraints, that a classical optimizer executes \citep{voxposer2023,rekep2024,copa2024}. The visual-interface robot agent of \citet{via2026} belongs to none of these: it exposes the robot as a visual interface and lets a general coding agent operate it through GUI tools with no robot-specific training. Harness VLA and ART likewise wrap or extend a VLA with agentic tool use \citep{harnessvla2026,art2026}, and Guava maps the design space of such harnesses (agent workflow, action space, observation space) and distills the result into a 4B model \citep{guava2026}. Since GPT-6 Astra, frontier models have also been benchmarked directly as robot policies, from tabletop manipulation to dexterous hands, navigation, and humanoid control \citep{galbot2026}. In all of these systems the harness around the task agent is designed by hand; it is the object we evolve.

\paragraph{Automatic harness optimization.}
For a fixed model, the harness largely determines agent performance: harness changes alone shift held-out success by 10 to 15 points on tool-use benchmarks \citep{beyondprompts2026}, an automatically discovered harness can outrank hand-built agents on Terminal-Bench \citep{metaharness2026}, and the best harness is model-specific because different models fail in different ways \citep{selfharness2026}. This has motivated optimizer agents that revise a harness from execution feedback. Meta-Harness gives the optimizer access to the code, scores, and traces of prior candidates \citep{metaharness2026}; Self-Harness lets the task agent mine its own failure clusters and propose minimal, regression-tested edits \citep{selfharness2026}; RHI iterates on prompt-level agent loops with pairwise feedback \citep{rhi2026}; and PRISM searches over prompt and middleware edits under a budget \citep{beyondprompts2026}. Closest to us, RHO brings harness search to robotics: a coding agent proposes and searches, at training time, a multi-file repository of prompts, tools, and control code that the deployed agent then runs, raising held-out success on an LLM-in-the-loop benchmark from 23.5\% to 44.3\% \citep{rho2026}. These methods treat the number of rollouts behind each decision and the rule that accepts a revision as fixed parts of the procedure rather than as variables under study. Recent critiques show that reported gains from harness evolution often disappear under matched budgets and held-out tasks \citep{rethinkingharness2026}, and that most claimed improvements on LIBERO are not statistically significant at the standard protocol \citep{benchmarking2026}. We bring the optimizer-agent loop to a visual-interface robot agent whose rollouts are long and noisy, use a strict train/test split, and treat the rollout scale behind each decision and the rule that selects revisions as the variables under study.

\paragraph{Agentic self-improvement in robotics.}
Coding agents have recently been used to improve robot systems autonomously. ASPIRE has an agent diagnose failures from execution traces, repair control programs, and accumulate a reusable skill library \citep{aspire2026}; ENPIRE automates the reset, train, roll out, and analyze loop for real-robot policy learning \citep{enpire2026}. These systems evolve skill code or a training pipeline, and the agent that performs the task is not itself a GUI-operating foundation-model agent. We instead evolve the harness of such a task agent, and ask what rollout scale and what selection rule the loop needs to improve reliably.

\section{Method}\label{sec:method}

\subsection{Setting}
We study harness evolution for the visual-interface robot agent of \citet{via2026}. The \emph{task agent} is an unmodified coding agent (Codex with GPT-5.6 Sol) that controls a simulated manipulator by taking screenshots of a browser-based 3D interface and calling a small set of MCP tools to pose a virtual target gripper, execute waypoints, and end the episode. Everything between the model and the robot constitutes the \emph{harness}: the system prompt and operating guide, the MCP tool set and the semantics of each tool (for example the maximum translation per call), the observation returned after each tool call, the control rules for staging and executing waypoints, and the verification and termination logic. All of this is ordinary code and text in the agent's repository and is the object of optimization. The task-agent model, the simulator, the task set, the success detectors, and the evaluation control plane are frozen and never edited.

\subsection{The harness-evolution loop}
Algorithm~\ref{alg:loop} summarizes the loop and Figure~\ref{fig:overview}(b) illustrates one round. We write $D$ for the fixed training sample of $B$ cases, $A$ and $O$ for the task agent and the optimizer agent, $\textsc{Evaluate}(A, H, D)$ for running $A$ under harness $H$ on every case in $D$, which returns the success count $s$ and the set of rollout records $R$, $\textsc{Evidence}(R)$ for the manifest $M$ built from those records, and $\mathcal{L}$ for the log of rejected challengers that is shown to the optimizer agent. The loop maintains a \emph{champion} harness $H_t$, initialized to the hand-written baseline harness $H_0$, and a fixed training sample of $B$ rollouts defined by a fixed set of tasks and seeds; the same $B$ cases are used in every round of a campaign.

\paragraph{Evaluate.} The champion is run on the $B$ cases. Each case is a full agentic rollout of the task agent, capped at 30 minutes of wall-clock time, and yields a binary success verdict; the success count $s(H_t)$, written $s$ in Algorithm~\ref{alg:loop}, is the champion's training score.

\paragraph{Collect evidence.} From the rollouts we build an \emph{evidence manifest} covering all $B$ cases: for each case, the ordered timeline of model turns, tool calls, controller feedback, and errors, together with the verdict and runtime statistics. The manifest retains the complete trajectory evidence of every case, so the optimizer agent reasons from full execution histories rather than from summary statistics.

\paragraph{Propose.} The \emph{optimizer agent} (also Codex with GPT-5.6 Sol, in a separate persistent session) reads the manifest and is instructed to identify one systemic, task-general failure mechanism and implement one coherent change to the harness. The change must be a single commit on top of $H_t$, must leave the evaluation control plane untouched, and must pass the repository's test suite; a candidate that violates any of these is returned for repair. The result is a \emph{challenger} harness $H'_t$. The optimizer agent is also shown the reasons for recently rejected challengers so that it does not repeat a regressive mechanism.

\paragraph{Select.} The challenger is evaluated on the same $B$ cases to obtain $s(H'_t)$, and a selection rule decides whether $H_{t+1} = H'_t$ or $H_{t+1} = H_t$.

A campaign runs this loop for a fixed number of rounds $T$ (30 in all main experiments). Held-out performance is measured only after the campaign, on a disjoint set of tasks, and is never visible to the optimizer agent.

\begin{algorithm}[t]
\caption{Harness evolution with Champion--Challenger selection}
\label{alg:loop}
\KwIn{baseline harness $H_0$; task agent $A$; optimizer agent $O$; fixed training sample $D$ of $B$ cases; number of rounds $T$}
\KwOut{champion harness $H_T$}
$H \leftarrow H_0$\; $(s, R) \leftarrow \textsc{Evaluate}(A, H, D)$ \tcp*{successes and rollouts on the fixed sample}
$\mathcal{L} \leftarrow \emptyset$ \tcp*{log of rejected challengers}
\For{$t \leftarrow 1$ \KwTo $T$}{
  $M \leftarrow \textsc{Evidence}(R)$ \tcp*{full trajectories and verdicts of all $B$ cases}
  $H' \leftarrow O(H, M, \mathcal{L})$ \tcp*{one bounded commit on top of $H$ that passes the gate}
  $(s', R') \leftarrow \textsc{Evaluate}(A, H', D)$ \tcp*{same $B$ cases}
  \eIf{$s' > s$}{
    $H \leftarrow H'$\; $(s, R) \leftarrow (s', R')$ \tcp*{promote the challenger}
  }{
    $\mathcal{L} \leftarrow \mathcal{L} \cup \{(H', s')\}$ \tcp*{reject; the champion is unchanged}
  }
}
\Return $H$ \tcp*{evaluated once on held-out tasks}
\end{algorithm}

\subsection{Rollout scale: the batch size \texorpdfstring{$B$}{B}}
$B$ is the central experimental variable. It controls how much information the optimizer agent receives each round: the trajectories it reads when diagnosing failures, and the outcomes on which each promotion is decided. Scaling laws in language modeling show that, with the model class held fixed, performance improves predictably as the amount of data seen during training grows. We ask whether an analogous regularity holds for harness evolution: with the optimizer agent, the task-agent model, the number of rounds, and the held-out set all held fixed, does the evolved harness improve as the number of rollouts per round grows? We therefore run otherwise identical campaigns with $B \in \{5, 10, 20, 100\}$. Each campaign has its own optimizer session, branch, and result namespace, so campaigns share no information. Because every round evaluates $B$ rollouts of both the champion and the challenger, $B$ also sets the cost of a campaign; the question is therefore equally whether spending more rollouts per round buys a better harness.

\subsection{Selection rule: unconditional acceptance versus Champion--Challenger}
The selection step admits two rules. Under \emph{unconditional acceptance}, $H_{t+1} = H'_t$ always: every revision the optimizer agent proposes becomes the base for the next round, and the harness follows a single linear chain of commits. Under \emph{Champion--Challenger} selection, the challenger is promoted if and only if
\begin{equation}
  s(H'_t) > s(H_t),
\end{equation}
that is, if it strictly beats the incumbent on the same cases. Ties and regressions are rejected; the rejected challenger and its evaluation are retained for audit, and the next round starts again from the unchanged champion. Section~\ref{sec:experiments} compares the two rules at equal $B$ and shows that unconditional acceptance degrades within ten rounds while Champion--Challenger improves over thirty.

\section{Experiments}\label{sec:experiments}

\subsection{Setup}\label{sec:setup}
\paragraph{Tasks and splits.}
The training sample is drawn from ten tabletop manipulation tasks in robosuite and LIBERO \citep{libero2023}: lift, square, and stack from robosuite, and seven LIBERO tasks spanning the Object, Goal, Spatial, 90, and 10 suites. The held-out set is ten different LIBERO tasks from the Object, Spatial, Goal, and 90 suites; the two task sets are disjoint. Every task is instantiated with fixed seeds that determine the initial object placement, so a case (task, seed) is reproducible across rounds and across harness versions. The largest training batch uses ten seeds per task ($B=100$); the held-out set always uses ten seeds per task (100 rollouts). Success is decided by each task's built-in detector. Section~\ref{sec:exp-robocasa} additionally uses RoboCasa \citep{robocasa2024}, a kitchen benchmark with articulated appliances and a mobile manipulator: ten kitchen tasks for harness evolution, run under our own protocol (ten seeds per task, a one-hour wall-clock limit per rollout; stated in full in Section~\ref{sec:exp-robocasa}), and the 50-task RoboCasa365 leaderboard benchmark \citep{robocasa365_2026}, run under the official protocol, for distilling the harness's rollouts back into a standalone policy (Section~\ref{sec:exp-distill}) and for combining the evolved harness with the distilled policy (Section~\ref{sec:exp-combo}).

\paragraph{Agents and protocol.}
The task agent is Codex with GPT-5.6 Sol at the highest reasoning effort, run through MCP tools with a 30-minute wall-clock limit per rollout; its model and settings are identical for every harness and every evaluation. The optimizer agent is the same model in a separate persistent session; each campaign has its own session, branch, and result namespace. Rollouts of one evaluation run concurrently in isolated sandboxes, so a $B=100$ evaluation takes about one wall-clock episode. All main campaigns run $T=30$ rounds. Held-out performance is measured once per selected harness after its campaign has finished; the optimizer agent never sees held-out tasks or results.

\subsection{Training batch size}\label{sec:exp-scaling}
\begin{figure}[t]
\centering
\IfFileExists{fig_scaling.pdf}{\includegraphics[width=\textwidth]{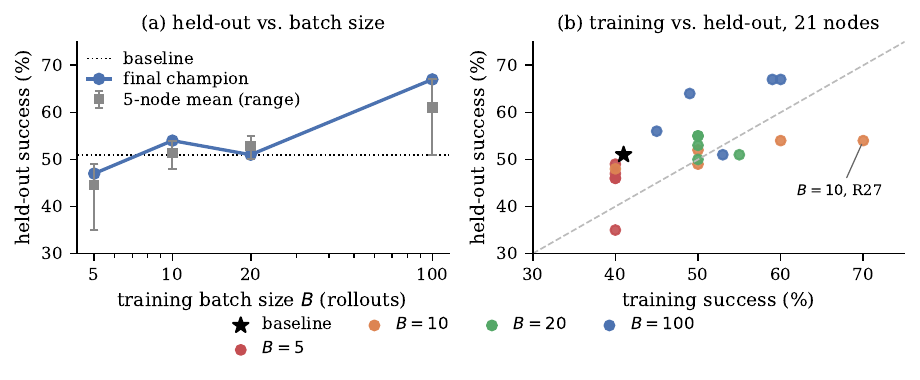}}{\fbox{\parbox{0.9\textwidth}{\centering\vspace{18mm}fig\_scaling.pdf (upload to project)\vspace{18mm}}}}
\caption{Training batch size. (a) Held-out success of the final champion of each 30-round campaign (blue) and the mean and range over the five held-out nodes of each campaign (grey), against the batch size $B$; the dotted line is the baseline harness. (b) Training success versus held-out success for the baseline and 20 evolved harnesses, coloured by $B$; points above the diagonal generalize beyond their training score. The $B=10$ champion with the highest training score of all campaigns, 70\%, reaches only 54\% held-out.}
\label{fig:scaling}
\end{figure}

\begin{table}[t]
\centering
\footnotesize
\setlength{\tabcolsep}{4pt}
\begin{tabular}{rcccc}
\toprule
$B$ & training success & promotions & held-out success & held-out success \\
 & baseline $\rightarrow$ final champion & (of 30) & final champion & 5-node mean (range) \\
\midrule
5 & 20\% $\rightarrow$ 40\% (R3) & 1 & 47\% & 44.6\% (35--49) \\
10 & 30\% $\rightarrow$ 70\% (R27) & 3 & 54\% & 51.4\% (48--54) \\
20 & 40\% $\rightarrow$ 55\% (R28) & 2 & 51\% & 52.8\% (50--55) \\
100 & 41\% $\rightarrow$ 60\% (R29) & 8 & \textbf{67\%} & \textbf{61.0\%} (51--67) \\
\midrule
\multicolumn{5}{l}{Baseline harness on the held-out set: 51\%.} \\
\bottomrule
\end{tabular}
\caption{Independent 30-round Champion--Challenger campaigns from the same baseline harness, differing only in the training batch size $B$. Held-out numbers are on 100 rollouts from tasks disjoint from training.}
\label{tab:scaling}
\end{table}

We run Champion--Challenger campaigns from the same baseline harness with $B \in \{5, 10, 20, 100\}$ and $T=30$, and evaluate the final champion of each on the held-out set. To see beyond the single final node, we also evaluate four further harnesses per campaign chosen from promoted and rejected challengers at different rounds, giving five held-out nodes per campaign and 21 nodes in total including the baseline. Table~\ref{tab:scaling} and Figure~\ref{fig:scaling} summarize the outcome.

\paragraph{Held-out success rises with $B$.}
The final champions reach 47\%, 54\%, 51\%, and 67\% held-out for $B = 5, 10, 20, 100$, against 51\% for the baseline harness (Figure~\ref{fig:scaling}a). The five-node means, 44.6\%, 51.4\%, 52.8\%, and 61.0\%, increase with $B$ and are less sensitive to which single round is picked. Only the $B=100$ campaign produces a clear gain: its final champion is 16 points above the baseline, and all five of its held-out nodes are at or above the baseline, three of them by 13 to 16 points. The $B=5$ campaign is the mirror image: all five of its nodes are below the baseline, one of them by 16 points, even though its final champion doubled its training score.

\paragraph{Small batches fit the training cases.}
Figure~\ref{fig:scaling}b plots each node's training score against its held-out score. The $B=100$ nodes lie above the diagonal: their held-out success exceeds their training success, and the two move together. The small-batch nodes scatter around or below it. The clearest case is the $B=10$ champion promoted at round 27 with a training score of 70\%, the highest of any campaign, which reaches 54\% held-out. Its training score is seven successes out of ten fixed cases; a harness change that helps two or three specific cases moves that score by 20 to 30 points while changing little on unseen tasks. Across all 21 nodes the correlation between training and held-out scores is 0.60, and it is carried by the $B=100$ nodes; within the small-batch campaigns, the training score does not rank the nodes correctly.

\paragraph{Larger batches also promote more.}
The number of promotions in 30 rounds is 1, 3, 2, and 8 for $B = 5, 10, 20, 100$ (Table~\ref{tab:scaling}). With five or ten cases most challengers tie the champion exactly and are rejected, so the campaign stalls; the $B=5$ campaign promoted once, at round 3, and rejected the remaining 27 challengers. With 100 cases a genuine improvement of a few percentage points registers as a strictly higher count, so the $B=100$ champion advanced eight times, at rounds 1, 7, 16, 18, 19, 23, 25, and 29, and was still improving when the campaign ended. Both effects, better generalization of what is promoted and more frequent promotion, follow from the optimizer agent seeing more rollouts per round.

\subsection{Selection rule}\label{sec:exp-selection}
\begin{figure}[t]
\centering
\IfFileExists{fig_selection_rule.pdf}{\includegraphics[width=\textwidth]{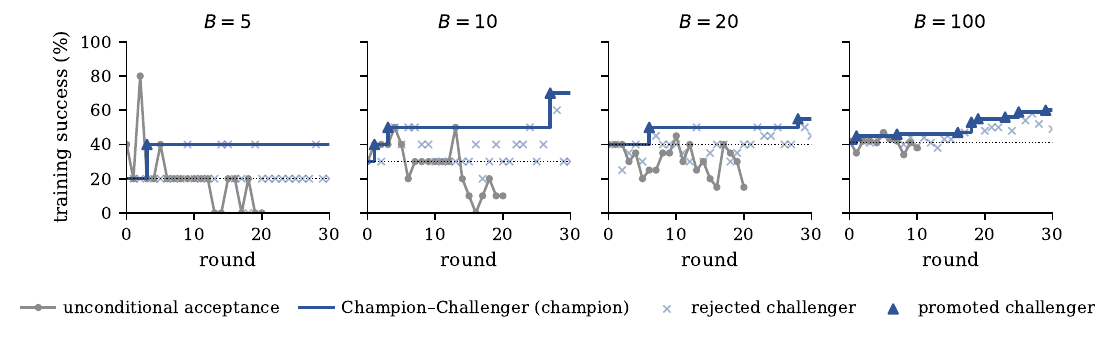}}{\fbox{\parbox{0.9\textwidth}{\centering\vspace{14mm}fig\_selection\_rule.pdf (upload to project)\vspace{14mm}}}}
\caption{Selection rule. Training success per round at each batch size under unconditional acceptance (grey; every revision becomes the next base) and Champion--Challenger selection (blue; the champion trajectory, with promoted challengers as triangles and rejected challengers as crosses). The dotted line is the baseline score. The unconditional runs were stopped after 10 rounds ($B=100$) or 20 rounds ($B=5,10,20$) once the downward trend was clear; their baseline commit was evaluated separately, so starting points differ slightly from the Champion--Challenger baselines.}
\label{fig:selection}
\end{figure}

To isolate the selection rule we compare, at each batch size, a campaign that accepts every revision unconditionally with the Champion--Challenger campaign of Section~\ref{sec:exp-scaling}. The unconditional campaigns use the same optimizer agent, task agent, evidence manifest, and fixed training samples; the only difference is that the challenger always becomes the next base. Figure~\ref{fig:selection} shows the training trajectories.

\paragraph{Unconditional acceptance drifts downward.}
Under unconditional acceptance, training success rises for a few rounds and then falls at every batch size. At $B=100$ the harness goes from 42\% to a peak of 47\% at round 5 and then to 34\% by round 8 and 38\% at round 10. At $B=20$ the harness falls from 40\% to 15\% within 20 rounds, at $B=10$ from 30\% to 10\%, and at $B=5$ from 40\% to 0\%. The mechanism is visible in the trajectories: a revision that scores well on one batch is kept, every later revision is built on top of it, and there is no way back when the next batch shows it was not an improvement. The harness simply accumulates every ill-judged edit.

\paragraph{Champion--Challenger turns the same loop into a monotone one.}
With the selection rule, the champion's training score is non-decreasing by construction, and the question is whether it keeps rising rather than stalling. At $B=100$ it does: from 41\% at the start to 60\% at round 29, with promotions spread across the 30 rounds. The rejected challengers, plotted as crosses, show what the rule is filtering out: 22 of 30 challengers at $B=100$ scored at or below the champion, and many scored well below it, so most of what the optimizer agent proposes would have been a regression. On the held-out set the final champion reaches 67\% against the baseline's 51\%, so the gains the rule admitted on the training sample are real improvements rather than fits to the fixed cases. Over the 20 evolved nodes, promoted harnesses average 54.0\% held-out and rejected ones 50.9\%.

\subsection{Transfer to a kitchen environment: RoboCasa}\label{sec:exp-robocasa}
\begin{figure}[t]
\centering
\IfFileExists{fig_robocasa.pdf}{\includegraphics[width=\textwidth]{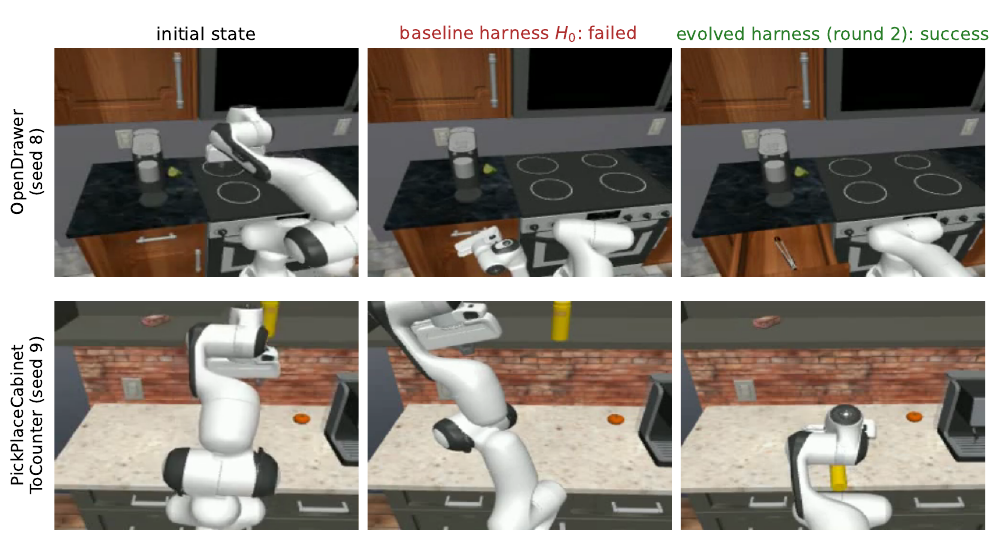}}{\fbox{\parbox{0.9\textwidth}{\centering\vspace{18mm}fig\_robocasa.pdf (upload to project)\vspace{18mm}}}}
\caption{Before and after harness evolution on RoboCasa. Two kitchen tasks with the same fixed seed, shown from the same camera: the initial scene, the final frame under the baseline harness, and the final frame under the round-2 champion. With the baseline the drawer stays shut and the bottle stays on the cabinet shelf; with the evolved harness the drawer is pulled open and the bottle is placed on the counter.}
\label{fig:robocasa}
\end{figure}
To check that the loop is not specific to the tabletop tasks above, we ran a four-round campaign on RoboCasa \citep{robocasa2024}, a kitchen benchmark with articulated appliances unlike the tabletop tasks above: ten kitchen tasks (drawers, a cabinet, a microwave, pick-and-place between counter, cabinet, sink, and stove, a stove knob, and a faucet) with ten seeds each, so $B=100$, and the same task agent. This campaign has no held-out set, so the scores are on the training sample.

\paragraph{Protocol.}
The kitchen numbers in this subsection are not leaderboard numbers and are not comparable with them. The ten tasks are a small subset of RoboCasa's kitchen tasks chosen for harness evolution, not the leaderboard's task set. Each task is run with ten seeds of our own under the simulator's default scene and object sampling rather than the benchmark's pretraining split, and a rollout counts as a success only if the task's built-in detector fires within a wall-clock limit of one hour and a cap of 4,500 simulator steps, the same cap for every task. The wall-clock limit is the binding constraint for an agentic rollout, in which every model turn takes seconds to minutes, so a harness that reaches the goal only slowly is scored as a failure. The leaderboard protocol instead fixes the object split, the seeds, and a per-task simulator-step horizon with no wall-clock limit; it is used only in Sections~\ref{sec:exp-distill} and \ref{sec:exp-combo}. Within this subsection, all harnesses and models are scored on the same 100 cases under the same limits, so the counts are comparable with one another.

\begin{table}[!htb]
\centering
\footnotesize
\begin{tabular}{lccl}
\toprule
round & harness & training success & decision \\
\midrule
0 & baseline harness & 2/100 & --- \\
1 & challenger & 26/100 & promoted \\
2 & challenger & \textbf{37/100} & promoted (final champion) \\
3 & challenger & 28/100 & rejected \\
4 & challenger & 33/100 & rejected \\
\bottomrule
\end{tabular}
\caption{Champion--Challenger campaign on ten RoboCasa kitchen tasks with $B=100$, scored under our protocol (ten seeds per task, success within a one-hour wall-clock limit and 4,500 simulator steps); not comparable with leaderboard numbers.}
\label{tab:robocasa}
\end{table}

The baseline harness, tuned on tabletop tasks, fails almost completely in the kitchen, succeeding in 2 of 100 rollouts. Two promoted revisions raised training success to 26\% and then 37\% (Table~\ref{tab:robocasa}, Figure~\ref{fig:robocasa}); the next two challengers scored 28\% and 33\% and were rejected, so the champion stayed at round 2. The same loop, with the same rollout scale and the same selection rule, therefore carries over to an environment it was not developed on: it finds large improvements on the optimization sample within two rounds, and the selection rule keeps the two later regressions out of the lineage.

\paragraph{The evolved harness transfers to a stronger model.}
The RoboCasa campaign evolved the harness with GPT-5.6 Sol as the task agent. To ask whether its gains are tied to that model, we re-ran the baseline harness and the round-2 champion on the same 100 cases, under the same protocol, with GPT-6 Astra, a newer and stronger model that the optimizer never saw (Table~\ref{tab:robocasa-astra}). The stronger model alone lifts the baseline harness from 2 to 37 successes; the evolved harness alone lifts Sol from 2 to 37; together they reach 83. The harness gain is not absorbed by the stronger model: under Astra it is worth 46 points, more than under Sol.

\begin{table}[!htb]
\centering
\footnotesize
\setlength{\tabcolsep}{5pt}
\begin{tabular}{lcccc}
\toprule
& \multicolumn{2}{c}{baseline harness $H_0$} & \multicolumn{2}{c}{evolved harness (round 2)} \\
\cmidrule(lr){2-3}\cmidrule(lr){4-5}
task agent & GPT-5.6 Sol & GPT-6 Astra & GPT-5.6 Sol & GPT-6 Astra \\
\midrule
successes / 100 & 2 & 37 & 37 & \textbf{83} \\
\bottomrule
\end{tabular}
\caption{Harness $\times$ model on the ten RoboCasa tasks. The harness was evolved with GPT-5.6 Sol only; GPT-6 Astra was never used during evolution. Successes on the same 100 cases (ten tasks, ten seeds each) under the protocol of this subsection: one-hour wall-clock limit per rollout, 4,500 simulator steps, our own seeds.}
\label{tab:robocasa-astra}
\end{table}

\subsection{From the evolved harness to a stronger policy}\label{sec:exp-distill}
The evolved harness is also a natural place to attach a learned policy, and, once attached, a data engine for improving that policy. Starting from the round-2 champion, we added one MCP tool, \texttt{vla\_run}, that hands control to the Xiaomi-Robotics-1 vision-language-action policy \citep{xiaomi2026} for a bounded number of simulator steps and returns a screenshot and the resulting gripper pose; the task agent (GPT-6 Astra) issues the instruction, judges the outcome from the screenshot, and either re-invokes the policy with a revised instruction or takes over with the original waypoint tools. Every rollout of this system is a trajectory in the policy's own action space, with the harness's decomposition, object choices, and take-overs written into the actions and labelled by the simulator's success detector. We first ask whether these rollouts can be written back into the policy, so that a single standalone policy, with no harness at inference, inherits part of what the harness adds; Section~\ref{sec:exp-combo} then puts the evolved harness on top of that stronger policy.

\begin{figure}[t]
\centering
\resizebox{\textwidth}{!}{%
\begin{tikzpicture}[
  x=1mm, y=1mm,
  font=\footnotesize,
  box/.style={draw, rounded corners=1pt, align=center, inner sep=2.5pt, minimum height=9mm, font=\footnotesize},
  agent/.style={box, fill=blue!10, thick},
  data/.style={box, fill=gray!12},
  hl/.style={box, fill=orange!18, thick, draw=orange!70!black},
  arr/.style={-{Latex[length=1.8mm]}, thick},
  lbl/.style={font=\scriptsize, fill=white, inner sep=1pt}
]
\node[hl, minimum width=28mm] (h) at (0,0) {evolved harness\\+ policy $\pi_0$};
\node[data, minimum width=28mm] (r) at (38,0) {rollouts at scale\\trajectories + verdicts};
\node[agent, minimum width=28mm] (e) at (76,0) {domain experts $\phi_1,\dots,\phi_K$\\one per task family};
\node[agent, minimum width=28mm] (m) at (114,0) {MOPD\\student trained on\\its own rollouts};
\node[hl, minimum width=28mm] (p) at (152,0) {distilled policy $\pi_1$\\no harness at test time};
\draw[arr] (h) -- (r);
\draw[arr] (r) -- (e);
\draw[arr] (e) -- (m);
\draw[arr] (m) -- (p);
\draw[arr] (p.south) -- ++(0,-8) -| (h.south);
\node[lbl] at (76,-8) {next cycle: $\pi_1$ replaces $\pi_0$ as the executor and harness evolution resumes on top of it};
\end{tikzpicture}%
}
\caption{Writing the rollouts back into the policy. The evolved harness drives the policy at scale; the recorded trajectories are written back as domain experts, one per task family; multi-teacher on-policy distillation (MOPD) merges the experts into one student trained on its own rollouts; the distilled policy runs without the harness. Section~\ref{sec:exp-combo} then evaluates the evolved harness on top of the distilled policy, which closes the loop.}
\label{fig:distill}
\end{figure}

\paragraph{Writing the rollouts back into the policy.}
The procedure has three stages, all in simulation and none using additional human demonstrations (Figure~\ref{fig:distill}). (i) \emph{Collect.} The evolved harness drives the policy on a large batch of rollouts, exactly as in the evolution loop, and the full trajectories are recorded together with their verdicts. (ii) \emph{Domain experts.} The rollouts are grouped by task family, and for each family a copy of the policy is fine-tuned on that family's successful trajectories mixed with the policy's original pretraining demonstrations. Each expert is specialized to its family and is not meant to be deployed on its own. (iii) \emph{Multi-teacher on-policy distillation.} A single student $\pi_\theta$, initialized from the original policy, is trained on its \emph{own} rollouts. Each rollout is assigned the expert $\phi_{k(o)}$ of its task family, and the student's flow-matching head is regressed onto the expert's velocity field evaluated at the student's own noisy-action samples under the expert's conditioning,
\begin{equation}
\mathcal{L}_{\mathrm{MOPD}} = \mathbb{E}_{o,\; a \sim \pi_\theta,\; t,\; \epsilon}\Big[\big\| v_\theta(x_t, t \mid o) - v_{\phi_{k(o)}}(x_t, t \mid o) \big\|^2\Big], \qquad x_t = (1-t)\,\epsilon + t\,a ,
\end{equation}
where $o$ is the observation, $a$ the action chunk the student executed, $\epsilon$ Gaussian noise, and $t$ the flow time. Because the targets are evaluated at the student's own samples, the objective is mode-seeking, the analogue of the reverse-KL objective used for on-policy distillation of language models, and the student's rollouts are used whether or not they succeed: failed rollouts are exactly the states on which the expert's correction matters. Human pretraining demonstrations are kept as a replay group with the standard flow-matching target. The experts are needed only during training; the student keeps the policy's original architecture and inference code. The distillation stage itself is short, 800 optimizer steps at an effective batch size of 256.

\paragraph{A stronger standalone policy.}
We evaluate the distilled policy under the official RoboCasa365 protocol (RoboCasa 1.0.1, pretraining split, official seeds, 50 tasks $\times$ 50 episodes; 2,497 of the 2,500 episodes completed, the remaining three excluded from the denominator) with no harness at inference (Table~\ref{tab:distill}). It reaches 62.2\% overall, 4.8 points above the policy it started from (57.4\%). The gain is concentrated on the composite-unseen split, 45.3\% against 32.1\%, while the two seen splits change by less than a point: what the written-back rollouts carry into the weights is above all the harness's handling of tasks the policy was not trained on, its object choices, task decomposition, and repositioning. The distilled policy has been submitted to the RoboCasa365 leaderboard (entry \emph{minionx}); verification by the benchmark maintainers was pending at the time of writing.

\begin{table}[!htb]
\centering
\footnotesize
\setlength{\tabcolsep}{3pt}
\begin{tabular}{lcccc}
\toprule
 & atomic-seen (18) & composite-seen (16) & composite-unseen (16) & overall (50) \\
\midrule
Xiaomi-Robotics-1 (policy alone) & 80.2 & 57.1 & 32.1 & 57.4 \\
distilled policy (MOPD, ours; no harness) & \textbf{80.9} & \textbf{58.0} & \textbf{45.3} & \textbf{62.2} \\
\bottomrule
\end{tabular}
\caption{A standalone policy distilled from the harness's rollouts: success rate (\%) on the RoboCasa365 benchmark, evaluated under the official protocol (50 tasks $\times$ 50 episodes, 2,497 episodes completed) with no harness at inference.}
\label{tab:distill}
\end{table}

\subsection{Combining the evolved harness with the distilled policy}\label{sec:exp-combo}
We now put the evolved harness on top of the distilled policy: \texttt{vla\_run} hands control to $\pi_1$ instead of the original policy, and everything else is unchanged. On the RoboCasa365 leaderboard benchmark (Table~\ref{tab:robocasa365}) the combination reaches 71.2\% overall, 13.1 points above the best published entry, Paimon-0 (58.1\%), 13.8 points above the original policy alone (57.4\%), and 9.0 points above the distilled policy alone (62.2\%).

Relative to the original policy the gains are nearly uniform, 13.5 to 14.1 points on each split, but Table~\ref{tab:robocasa365} shows that they arrive in two different ways. On the composite-unseen split the distilled policy alone already reaches 45.3\%, and the harness on top of it adds only 0.3 points: the object choices, task decomposition, and repositioning that the agent supplied on unseen tasks have been written into the weights. On the two seen splits the distilled policy is within a point of the original, and the harness adds 13.0 and 13.2 points at test time; on these splits the policy does almost all the work, with 95\% of successful episodes finishing without the agent touching a waypoint tool, and the harness's contribution is monitoring the policy, re-issuing instructions, and taking over when it stalls. Overall, 4.8 of the 13.8 points over the original policy are carried by the distilled policy alone and 9.0 by the harness on top of it.

\begin{table}[!htb]
\centering
\footnotesize
\setlength{\tabcolsep}{3pt}
\begin{tabular}{lcccc}
\toprule
 & atomic-seen (18) & composite-seen (16) & composite-unseen (16) & overall (50) \\
\midrule
Paimon-0 (leaderboard \#1) & 79.3 & 55.8 & 36.7 & 58.1 \\
Xiaomi-Robotics-1 (policy alone) & 80.2 & 57.1 & 32.1 & 57.4 \\
distilled policy $\pi_1$ (MOPD, ours; no harness) & 80.9 & 58.0 & 45.3 & 62.2 \\
evolved harness + distilled policy $\pi_1$ (ours) & \textbf{93.9} & \textbf{71.2} & \textbf{45.6} & \textbf{71.2} \\
\bottomrule
\end{tabular}
\caption{Success rate (\%) on the RoboCasa365 leaderboard benchmark. The two bottom rows are ours (the distilled policy under the official protocol, Table~\ref{tab:distill}, and the evolved harness on top of it); the other rows are leaderboard results.}
\label{tab:robocasa365}
\end{table}

\paragraph{Closing the loop.}
The combination above is already the first turn of a loop: the harness's rollouts produced a stronger policy, and that policy now sits behind \texttt{vla\_run} as the executor. The loop closes when harness evolution resumes on top of $\pi_1$ and the new rollouts are written back again. Section~\ref{sec:exp-robocasa} showed that the evolved harness is not tied to the model it was evolved with, so a stronger executor raises the whole combination, which in turn produces better rollouts, better experts, and a better distilled policy. We have run one such cycle; the pipeline is built so that the cycle repeats, with the harness and the policy improving each other, and the cost of each cycle is simulation rollouts and one short distillation stage.

\section{Conclusion}\label{sec:conclusion}
Automatic harness evolution for a visual-interface robot agent needs enough rollouts behind each decision, which lift held-out success from 47\% at $B=5$ to 67\% at $B=100$, and Champion--Challenger selection, without which the loop degrades. Writing the harness's rollouts back into the policy gives a standalone policy at 62.2\% on RoboCasa365, and the harness on top of it reaches 71.2\%, 13.1 points above the best published entry, closing the loop.

\bibliographystyle{plainnat}
\bibliography{references}

\end{document}